\documentclass[letter,twocolumn]{jpsj3}
\usepackage{txfonts}
\usepackage{amsmath}
\usepackage{bm}
\usepackage{multirow}
\usepackage{url}
\usepackage{algorithmic}
\usepackage{algorithm}
\title{Statistics of Min-max Normalized Eigenvalues in Random Matrices}

\author{Hyakka Nakada$^{1}$\thanks{hyakka\_nakada@keio.jp} and Shu Tanaka$^{1,2,3,4,5}$}
\inst{$^1$Graduate School of Science and Technology, Keio University, Kanagawa 223-8522, Japan \\
$^2$Department of Applied Physics and Physico-Informatics, Keio University, Kanagawa 223-8522, Japan \\
$^3$Keio University Sustainable Quantum Artificial Intelligence Center (KSQAIC), Keio University, Tokyo 108-8345, Japan \\
$^4$Human Biology-Microbiome-Quantum Research Center (WPI-Bio2Q), Keio University, Tokyo 108-8345, Japan \\
$^5$Green Computing System Research Organization, Waseda University, Shinjuku, Tokyo 162-0042, Japan
}

\abst{Random matrix theory has played an important role in various areas of pure mathematics, mathematical physics, and machine learning.
From a practical perspective of data science, input data are usually normalized prior to processing. Thus, this study investigates the statistical properties of min-max normalized eigenvalues in random matrices.
Previously, the effective distribution for such normalized eigenvalues has been proposed.
In this study, we apply it to evaluate a scaling law of the cumulative distribution.
Furthermore, we derive the residual error that arises during matrix factorization of random matrices.
We conducted numerical experiments to verify these theoretical predictions.
}

\begin{document}
\maketitle

Practical data analysis pipelines often employ preprocessing steps. Among these techniques, feature scaling, which scales all features to a fixed range $[0, 1]$, is widely adopted for purposes such as mitigating the influence of extremely large values and stabilizing machine learning models.
Furthermore, this scaling facilitates interpretation as rates or probabilities. 
For example, in principal component analysis (PCA), the proportion of variance explained by each principal component is crucial for measuring its contribution, which corresponds to the relative magnitude of each eigenvalue~\cite{pca}.
Thus, normalized quantities or ratios can sometimes be better statistical measures than absolute values.
Representative examples of such scaling methods include min-max normalization and the softmax activation function.

Random matrix theory has played an important role in various areas of pure~\cite{rmt1,rmt2} and applied mathematics~\cite{rmt3,rmt4,rmt5,rmt6}.
For example, in physics, the interaction coefficients in spin glass models are typically described by random matrices~\cite{rmt_spin}.
In computer science, because real-world signals contain a significant amount of noise, random matrices are sometimes used to model the data matrix~\cite{rmt_data}.
A key advantage is that one can predict the asymptotic spectral properties even without knowing the detailed values of the matrix elements.
In recent years, the size of data matrices has grown enormously, leading to active study on random matrix theory~\cite{rmt_deep1,rmt_deep2,rmt_deep3}.

This paper investigates the theoretical properties of the distribution of min-max normalized eigenvalues in random matrices,
\begin{align}
\hat{\lambda} = \frac{\lambda-\lambda_N}{\lambda_1-\lambda_N}. \label{eq:1}
\end{align}
Here, $\lambda_1$ and $\lambda_N$ are the largest and smallest eigenvalues of a random matrix $Q$, which follows Gaussian distribution,
\begin{align}
Q_{ij} \overset{\textrm{iid}}{\sim} \mathcal{N}\left(\mu,\sigma^2\right) \label{eq:2}
\end{align}
where $i < j$. The mean of the coupling coefficient is denoted by $\mu$, the standard deviation by $\sigma^2$, and the input dimension is $N$. In addition, we assume that diagonal $Q_{ii}$ follows $N(\mu,2\sigma^2)$. 
Thus, the matrix $Q$ is an orthogonal random matrix with a finite mean.
Hereinafter, $\mu>0$ is considered.

In the previous study~\cite{rmt_fma}, we proposed the cumulative distribution of the min-max normalized eigenvalues~\eqref{eq:1}, during the process of analyzing Factorization Machines (FMs)~\cite{fm1,fm2}.
FMs are a supervised learning model that can estimate pairwise feature interactions by factorizing the interaction matrix into low-rank latent vectors, making them effective for tasks like recommendation in sparse datasets.
Recently, FMs have also been applied to black-box optimization~\cite{fma1,fma2,fma3,fma4,fma5,fma6,fma7}. Because these models are considered as Quadratic Unconstrained Binary Optimization (QUBO) forms, it has also garnered significant attention as an application of quantum annealing~\cite{qa1,qa2,qa3,qa4}.
By approximating the eigenvalues by their expectations as follows,
\begin{align}
\label{eq:3}
&\lambda_1 \approx E\left[\lambda_1\right] = N\mu+\sigma^2/\mu \nonumber \\
&\lambda_2 \approx E\left[\lambda_2\right] = 2\sqrt{N}\sigma \\ 
&\lambda_N \approx E\left[\lambda_N\right] = -2\sqrt{N}\sigma, \nonumber
\end{align}
we obtained the cumulative distribution~\cite{rmt_fma}:
For $\sigma\leq \sqrt{N} \mu$,
\begin{align}
P\left( \hat{\lambda}<x \right)=
\begin{cases}  
1+\frac{x-1}{(1-r)N} & \text{for $x>r$,} \\
\frac{1}{N}+\frac{4(N-2)}{\pi N r^2} \displaystyle\int_{0}^{x}\sqrt{r^2-(2t-r)^2}dt & \text{for $x\leq r$}.
\end{cases}
\label{eq:4}
\end{align}
For $\sigma > \sqrt{N} \mu$,
\begin{align}
P\left( \hat{\lambda}<x \right)=
\frac{1}{N}+\frac{4(N-1)}{\pi N} \displaystyle\int_{0}^{x}\sqrt{1-(2t-1)^2}dt.
\label{eq:5}
\end{align}
In Eq.~\eqref{eq:3}, the expectation values are derived from Wigner’s semicircle law~\cite{rmt_semicircle1, rmt_semmicircle2} and the theory of the largest eigenvalue distribution theory~\cite{rmt_largest1, rmt_largest2}.
$r$ denotes the deterministic value for the normalized second-largest eigenvalue $\hat{\lambda}_2$,
\begin{align}
r \equiv \frac{4\sqrt{N}\mu\sigma}{(\sqrt{N}\mu+\sigma)^2}.
\label{eq:6}
\end{align}

In this study, starting from the approximate cumulative-distribution~\eqref{eq:4}, \eqref{eq:5} and \eqref{eq:6}, several statistics are derived and discussed.
Firstly, we explain the nature of scaling laws. The behavior of normalized eigenvalues essentially depends solely on the ratio of standard deviation to mean, as detailed later.
Secondly, the residual error is analytically derived when $Q$ is decomposed by matrix factorization~\cite{mf}.
Matrix factorization has a wide variety of applications, such as dimensionality reduction~\cite{mf_dim}, recommender system~\cite{mf_rec}, and model compression~\cite{mf_comp}. This is also the foundation of FMs~\cite{fm2}.
We perform experiments on the consistency between the predicted error and the empirical error, to verify our theory.

Firstly, we discuss the scaling laws.
According to the basic treatment used in spin glass~\cite{spin_glass}, the mean and standard deviation are rewritten as
\begin{align}
&\mu= \frac{J_0}{N}, \sigma= \frac{J_1}{\sqrt{N}}.
\label{eq:7}
\end{align}
Here, the values of $J_0$ and $J_1$ are positive.
Equation~\eqref{eq:6} is transformed to
\begin{align}
\label{eq:8}
r=\frac{4J_1/J_0}{(1+J_1/J_0)^2}.
\end{align}
Thus, $r$ depends only on the value of $J_1/J_0$.
Equation~\eqref{eq:8} has the value of $1$ if and only if $J_0=J_1$.
When the input dimension $N$ is sufficiently large, Eqs.~\eqref{eq:4} and \eqref{eq:5} converge to the following distributions.
For $J_1 \leq J_0$,
\begin{align}
P\left( \hat{\lambda}<x \right)=
\begin{cases}  
1 & \text{for $x>r$,} \\
\frac{4}{\pi r^2} \displaystyle\int_{0}^{x}\sqrt{r^2-(2t-r)^2}dt & \text{for $x\leq r$}.
\end{cases}
\label{eq:9}
\end{align}
For $J_1 > J_0$,
\begin{align}
P\left( \hat{\lambda}<x \right)=
\frac{4}{\pi } \displaystyle\int_{0}^{x}\sqrt{1-(2t-1)^2}dt.
\label{eq:10}
\end{align}
Based on Eqs.~\eqref{eq:8}, \eqref{eq:9} and \eqref{eq:10}, the cumulative distribution of min-max normalized eigenvalues is expected to depend solely on the ratio of $J_1$ to $J_0$.  
This is a property not observed in the distribution of unnormalized eigenvalues.
\begin{figure}[t]
\centering
\includegraphics[width=8.5cm]{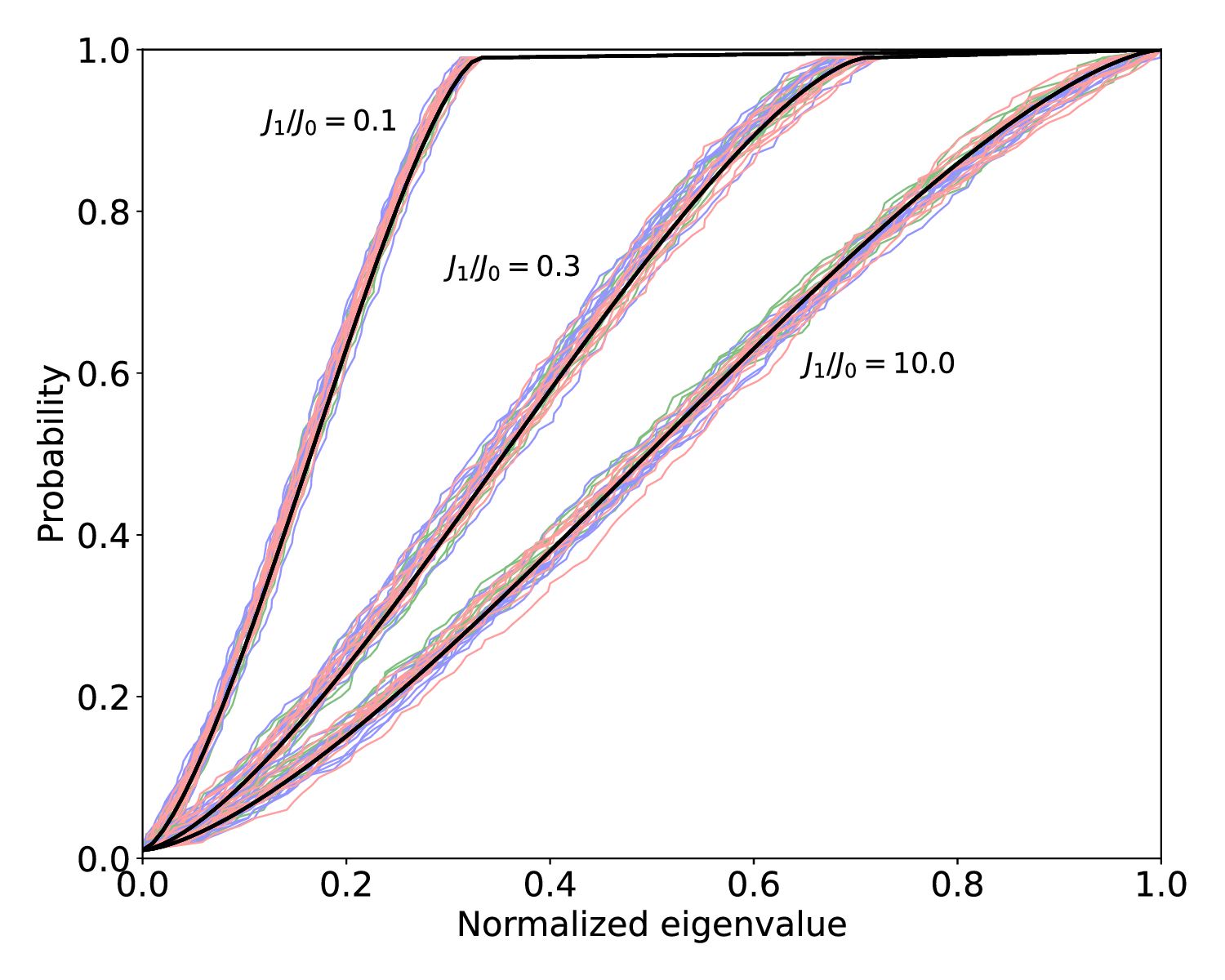}
\caption{(Color online) Plots of the cumulative distribution of min-max normalized eigenvalues with varying $J_1/J_0$. The black lines show our theoretical line computed from Eq.~\eqref{eq:4} or \eqref{eq:5}, for $J_1/J_0=0.1, 0.3, 10.0$ respectively. The green, blue, and red lines are experimental plots for $J_0 = 0.1, 1.0, 10.0$, respectively. These lines were obtained with $N=100$.}
\label{fig:result_scaling}
\end{figure}

\begin{figure}[thb]
\centering
\includegraphics[width=8.5cm]{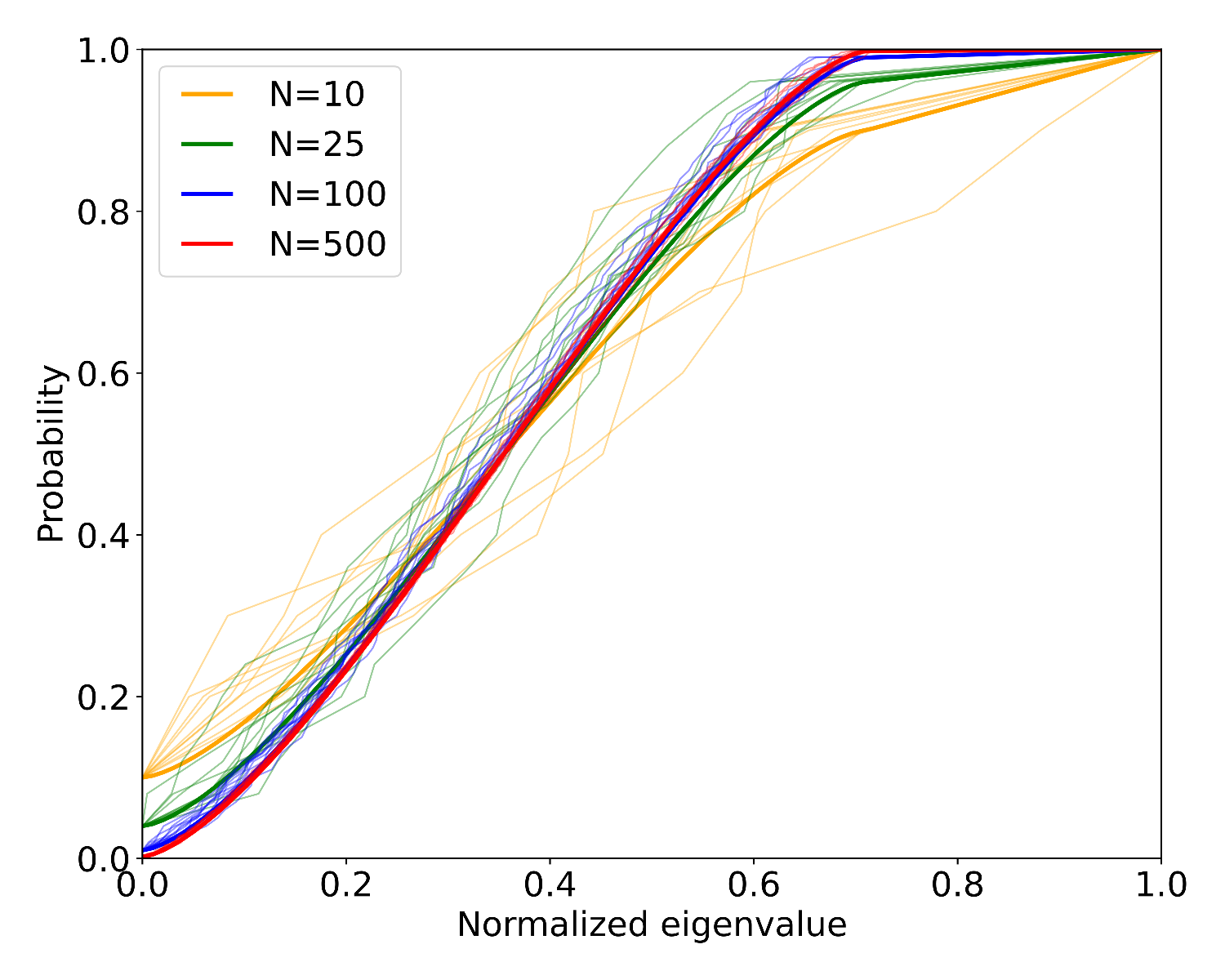}
\caption{(Color online) Plots of the cumulative distribution of min-max normalized eigenvalues with varing $N$. The yellow, green, blue, and red thick lines show our theoretical lines computed from Eq.~\eqref{eq:4}, for $N=10, 25, 100, 500$ respectively.
The thin lines represent the experimental results. These lines were obtained with $J_0=1$ and $J_1=0.3$.}
\label{fig:result_asymptonic}
\end{figure}

To verify the above scaling law, we performed the following numerical experiment. After sampling random matrices $Q$, eigenvalues were numerically calculated by eigenvalue decomposition. 
Then, the cumulative distribution was experimentally estimated.
For $J_0 = 0.1, 1.0, 10.0$, the values of $J_1$ were designed to keep $J_1/J_0=0.1$.
The plots of the cumulative distribution are shown in Fig.~\ref{fig:result_scaling}.
Here, green, blue, and red lines are the distributions for $J_0 = 0.1, 1.0, 10.0$, respectively. The theoretical line~\eqref{eq:4} is indicated by a black line.
Thus, the experimental result agrees with our theory well, regardless of the value of $J_0$.
These calculations were performed also for $J_1/J_0=0.3, 10.0$. 
Specifically for $J_1/J_0=10.0$, we adopted Eq.~\eqref{eq:5} as the theoretical line because the condition of $J_1>J_0$ holds. 
In all cases, the cumulative distributions of min-max normalized eigenvalues were found to depend only on $J_1/J_0$. 
Through this experiment, the input dimension $N$ was set to $100$, and $10$ trials were randomly performed for each parameter set of $(J_0, J_1)$.
Note that the plots start slightly above the origin because min-max normalization sets the minimum eigenvalue to zero, resulting in a non-zero probability at $x=0$.
In addition, the asymptotic behavior along $N$ was also checked.
The results are shown in Fig.~\ref{fig:result_asymptonic}.
The cumulative distributions were estimated for $N=10,25,100,500$.
From this figure, one can observe the convergence to a unique distribution, which corresponds to Eq.~\eqref{eq:9}. 
Through this experiment, $J_0$ and $J_1$ were fixed to $1$ and $0.3$, respectively, and $10$ independent trials were performed for each parameter set of $N$.
\begin{figure*}[tb]
  \begin{minipage}[b]{0.67\columnwidth}
    \centering
    \includegraphics[width=5.8cm]{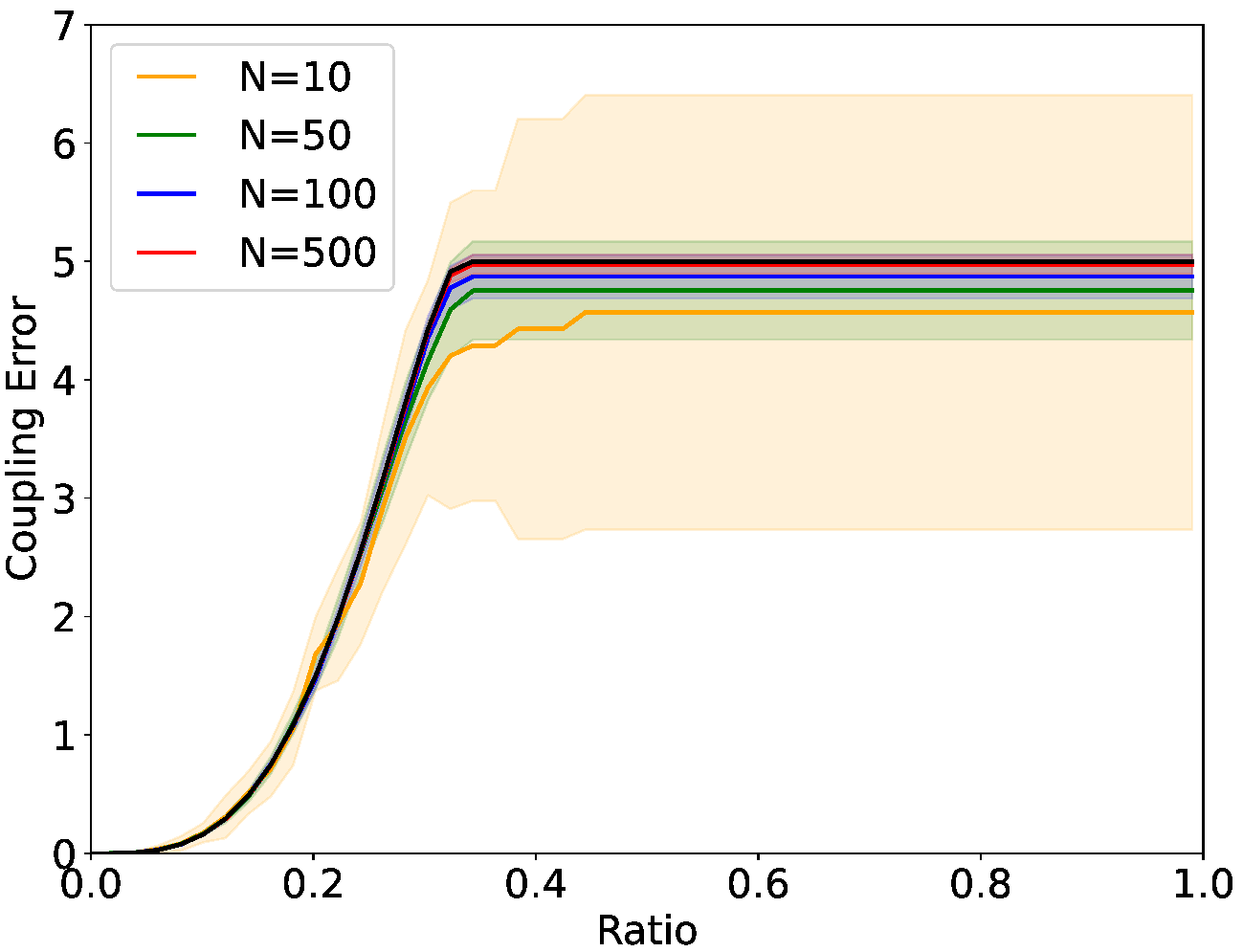}
  \end{minipage}
  \begin{minipage}[b]{0.67\columnwidth}
    \centering
    \includegraphics[width=5.8cm]{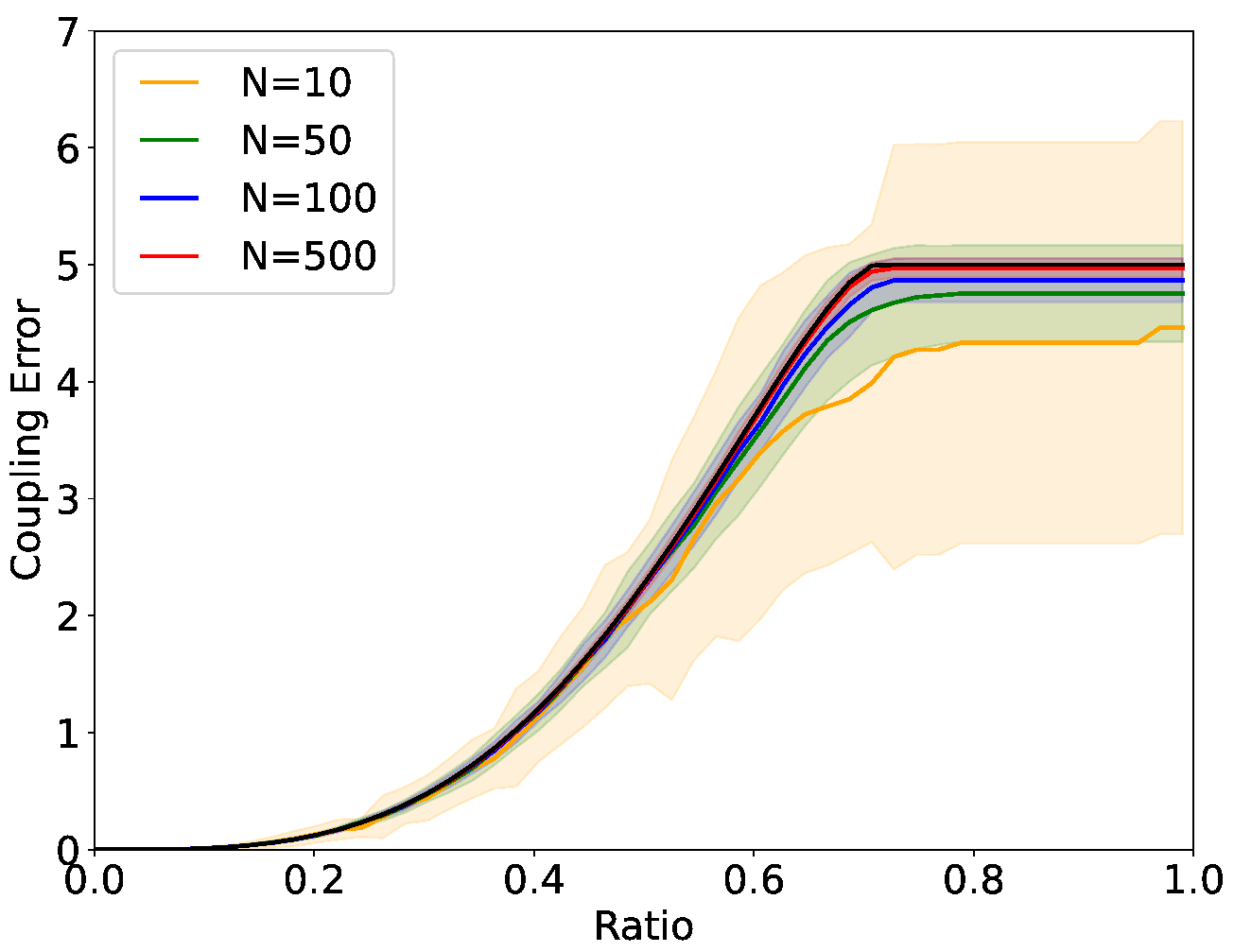}
  \end{minipage}
  \begin{minipage}[b]{0.67\columnwidth}
    \centering
    \includegraphics[width=5.8cm]{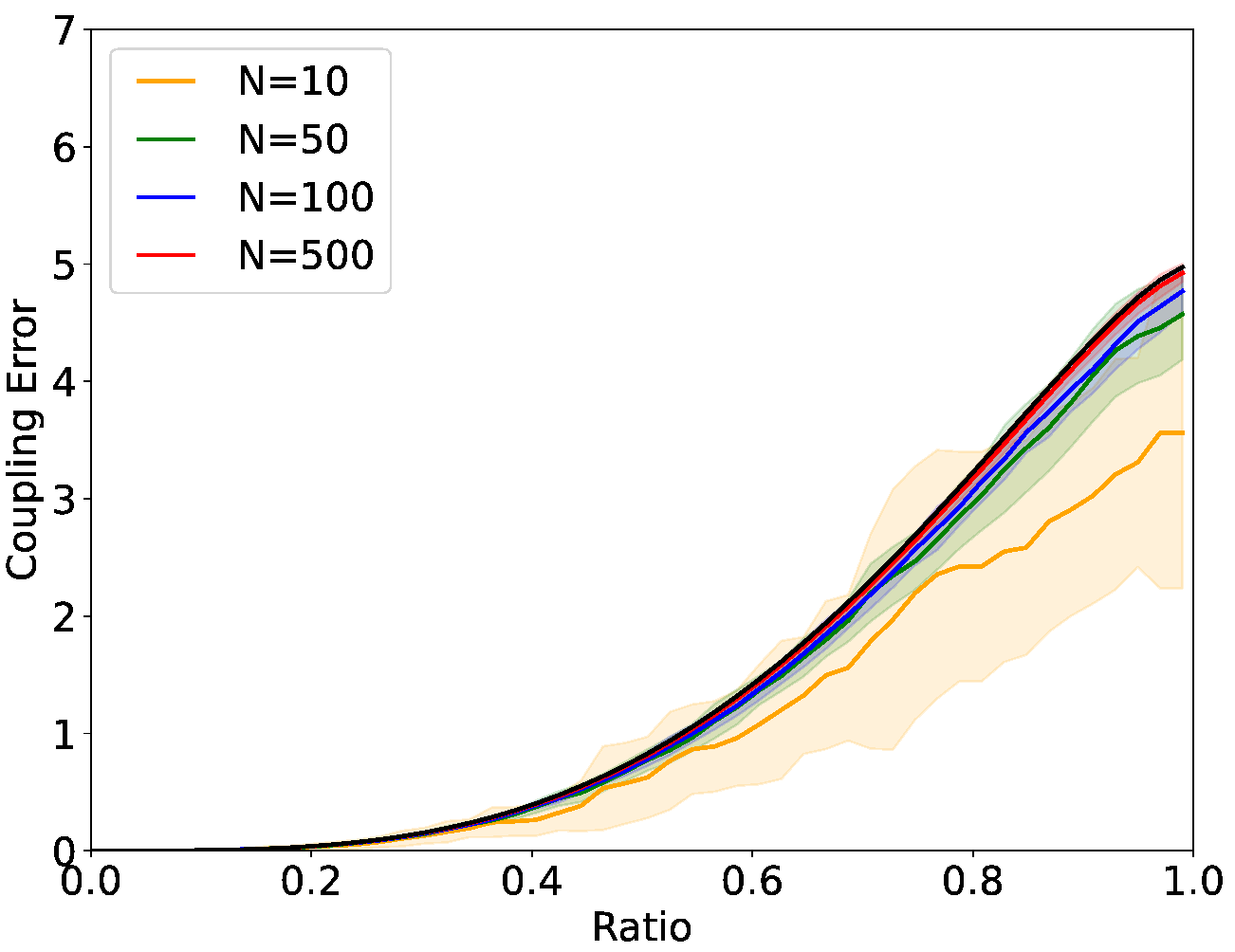}
  \end{minipage}
\caption{(Color online) Plots of coupling errors as a function of ratio $\alpha$. The black line shows our theoretical prediction computed from Eq.~\eqref{eq:16} or \eqref{eq:17}. The yellow, green, blue and red lines represent the experimental results for $\Delta_{\alpha}$.
These lines shall guide the eye. The colored areas indicate the standard deviations.
With fixing $J_0=1$, the results of $J_1=0.1$ (left), $0.3$ (middle), and $10.0$ (right) are shown.
}
\label{fig:result_1}
\end{figure*}

\begin{figure}[tbh]
\centering
\includegraphics[width=7cm]{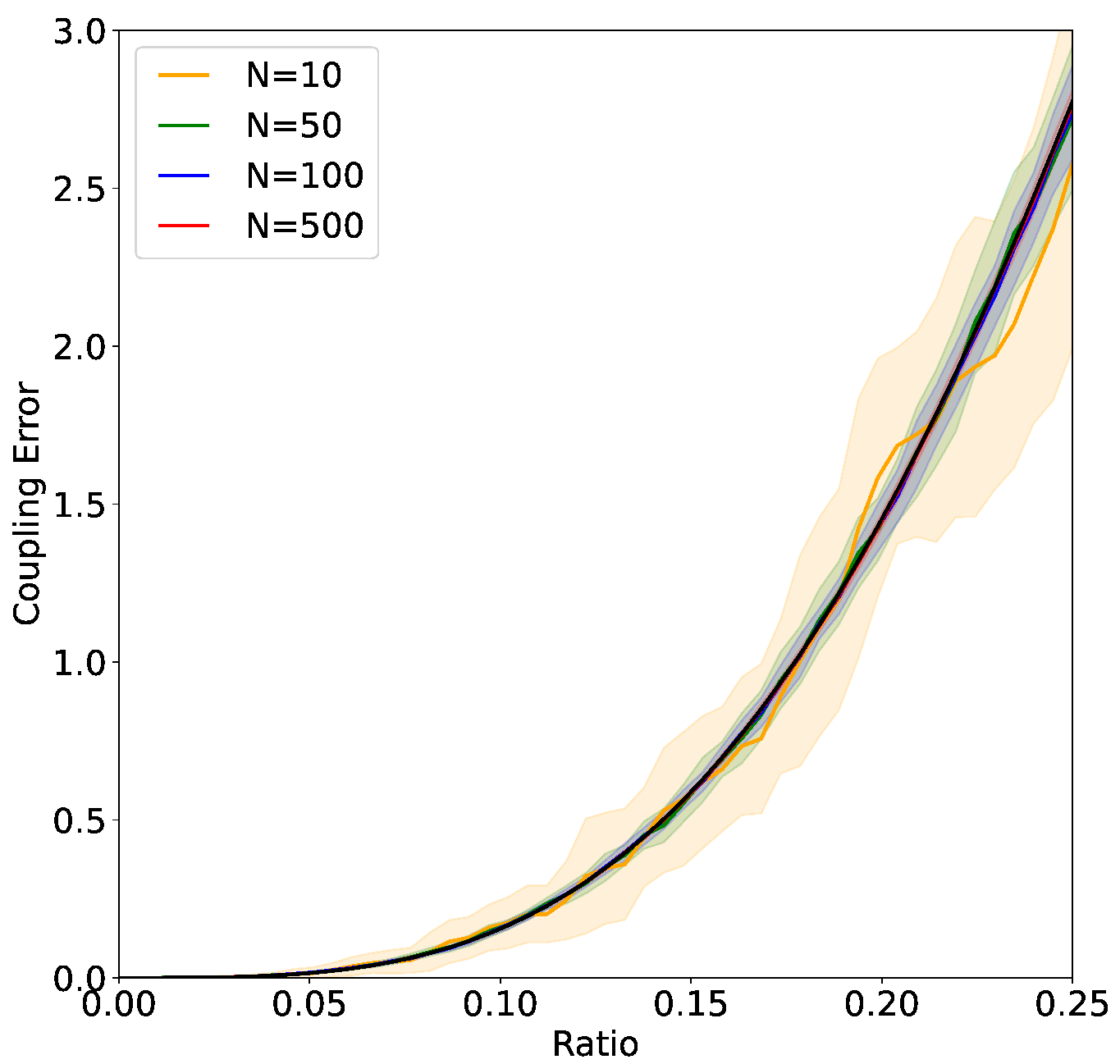}
\caption{(Color online) Cropped and enlarged view of Fig.~\ref{fig:result_1} at $J_1=0.1$.}
\label{fig:result_1_up}
\end{figure}

Secondly, we discuss the residual error on matrix factorization $V V^T$ of $Q$.
Such a factorization is justified when the target matrix is semi-definite.
Thus, we consider the factorization of a regularized matrix,
\begin{align}
Q-\lambda_N I \approx V V^T.
\label{eq:11}
\end{align}
$I$ is an identity matrix.
In matrix factorization, the fidelity and the generality are modulated by the rank $k$, which corresponds to the number of rows of $V\in \mathbb{R}^{N \times k}$.
$Q-\lambda_N I$ can be decomposed by eigenvalue decomposition,
\begin{align}
Q-\lambda_N I = U\Sigma U^T=\left(U\Sigma^{1/2}\right)\left(U\Sigma^{1/2}\right)^T.
\label{eq:12}
\end{align}
Here, $\Sigma$ is $\text{diag}(\lambda_1-\lambda_N,\lambda_2-\lambda_N,\ldots, \lambda_{N-1}-\lambda_N,0)$ and $U$ is a unitary matrix.
Because all diagonal components are nonnegative, $\Sigma^{1/2}$ is well-defined.
Thus, if $k$ is set to a full rank $N$, adopting $V=U\Sigma^{1/2}$ does not produce residual errors. 
In other words, the matrix is perfectly factorized.
Generally, reducing the rank $k$ leads to an increase in the error.

When the rank $k$ is lower than $N-1$, 
\begin{align}
Q-\lambda_N I \approx U'\Sigma' U'^T=\left(U'\Sigma'^{1/2}\right)\left(U'\Sigma'^{1/2}\right)^T
\label{eq:13}
\end{align}
is obtained.
Here, $\Sigma'$ is $\text{diag}(\lambda_1-\lambda_N,\ldots, \lambda_{k}-\lambda_N)$ and $U'\in \mathbb{R}^{N\times k}$ is a truncated matrix of $U$.
The residual error of Eq.~\eqref{eq:13} is calculated using the squared Frobenius norm,
\begin{align}
\Delta_{k} &= \left\|Q -\lambda_N I-U'\Sigma^{\prime}U'^T \right\|^2 \nonumber \\
&= \sum^{N}_{i=k+1} \left(\lambda_i-\lambda_N\right)^2 \nonumber \\
&=\sum^{N}_{i=k+1} \left((\lambda_1-\lambda_N)\hat{\lambda}_i\right)^2 \nonumber \\
&\approx \frac{16 J_1^2}{r^2}\sum^{N}_{i=k+1} \left(\hat{\lambda}_i\right)^2.
\label{eq:14}
\end{align}
From the third to the fourth term, we approximated $\lambda_1-\lambda_N$ by the deterministic value determined by Eq.~\eqref{eq:3}.
We call the residual error ``coupling error'' because $Q$ corresponds to the interaction matrix in the Sherrington--Kirkpatrick (SK) model~\cite{sk1,sk2}, according to the previous study~\cite{rmt_fma}.
Note that the error~\eqref{eq:14} is actually with respect to $Q-\lambda_N I$ rather than $Q$, but in Ising spin systems like the SK model, shifts in the diagonal elements do not change the physics. Thus, it can be considered an effective error for $Q$.

Our objective is to revisit Eq.~\eqref{eq:14} in terms of the ratio of eigenvalues.
In other words, we define it as a function of the threshold ratio $\alpha$ instead of the rank $k$: $\Delta_{\alpha} \propto \sum_{\hat{\lambda}_i<\alpha}(\hat{\lambda}_i)^2$. This represents the coupling error by truncating normalized eigenvalues less than $\alpha$.  
By considering the expectation using the continuum approximation,
\begin{align}
E\left[\Delta_{\alpha}\right] \approx \frac{16J_1^2}{r^2} \int^{\alpha}_0 x^2 NdP\left( \hat{\lambda}<x \right)
\label{eq:15}
\end{align}
Integrating Eq.~\eqref{eq:15} with the distributions~\eqref{eq:4} and \eqref{eq:5} yields the analytical forms of the expected coupling error.
For $J_1\leq J_0$, we obtain
\begin{align}
\frac{E\left[\Delta_{\alpha}\right]}{N J_1^2} &\approx
\begin{cases}
\frac{1}{3N}\left(15 (N-2)+16\frac{\alpha^3-r^3}{(1-r)r^2} \right)& \text{for $\alpha > r$,} \\
\frac{N-2}{3\pi N} g\left(\frac{\alpha}{r}\right) & \text{for $\alpha\leq r$,} 
\end{cases}
\label{eq:16}
\end{align}
for $J_1>J_0$,
\begin{align}
\frac{E\left[\Delta_{\alpha}\right]}{N J_1^2} &\approx
\frac{N-1}{3 \pi N} g(\alpha).
\label{eq:17}
\end{align}
Here, $g(x)=15(\pi-\arccos(2x-1)) -2\sqrt{x(1-x)}(15+10x+8x^2-48x^3)$.
In the case of sufficiently large $N$, Eqs.~\eqref{eq:16} and \eqref{eq:17} will converge to the following errors.
For $J_1\leq J_0$,
\begin{align}
\frac{E\left[\Delta_{\alpha}\right]}{N J_1^2} &\approx
\begin{cases}
5 & \text{for $\alpha > r$,} \\
\frac{1}{3\pi} g\left(\frac{\alpha}{r}\right) & \text{for $\alpha\leq r$,} 
\end{cases}
\label{eq:18}
\end{align}
for $J_1>J_0$,
\begin{align}
\frac{E\left[\Delta_{\alpha}\right]}{N J_1^2} &\approx
\frac{1}{3 \pi} g(\alpha).
\label{eq:19}
\end{align}
Thus, from the above equations~\eqref{eq:18} and \eqref{eq:19}, the coupling error normalized by $N J_1^2$ is expected to depend solely on the ratio of $J_1$ to $J_0$, in the same manner as the cumulative distributions~\eqref{eq:9} and \eqref{eq:10}.

To verify our theoretical coupling error, we performed the following numerical experiment. After sampling random matrices $Q$, we calculated the eigenvalues numerically via eigenvalue decomposition. By sorting the eigenvalues in descending order, we obtained $\lambda_i \ \text{for}\ i=1,\ldots,N$.
Then, the sum of squared values of $(\lambda_i-\lambda_N)$ was calculated for the components satisfying the truncation condition $\lambda_i-\lambda_N<\alpha (\lambda_1-\lambda_N)$. 
This is the numerical value of the coupling error $\Delta_\alpha$.
Figure~\ref{fig:result_1} shows the plots of $\Delta_\alpha$ with respect to the threshold ratio $\alpha$.
The solid lines represent the means of coupling errors at each $\alpha$, averaged over $20$ trials.
The colored areas around the solid lines denote their variances.

With the value of $J_0$ fixed to $1$, the theoretical line of Eq.~\eqref{eq:18} or \eqref{eq:19} is additionally shown in Fig.~\ref{fig:result_1}, at $J_1=0.1, 0.3, 10.0$ and $N=10, 50, 100, 500$.
Fixing $J_0$ is justified by the nature of Eqs.~\eqref{eq:9} and \eqref{eq:10}. In other words, the behavior of normalized eigenvalues depends only on $J_1/J_0$, as explained in the scaling law.
At $J_1=0.1, 0.3$, Eq.~\eqref{eq:18} was adopted for the theoretical line because the condition of $J_1 \leq J_0$ holds.
On the other hand, Eq.~\eqref{eq:19} was used at $J_1=10.0$.
Figure~\ref{fig:result_1_up} is the magnified view of Fig.~\ref{fig:result_1} at $J_1=0.1$, for visibility.
From these figures, convergence to the theoretical line can be seen with respect to the input dimension $N$. 
In addition, the theoretical lines are located almost in the region of variances.
Here, note that the standard deviation increases as the value of $N$ decreases due to $\sigma \propto 1/\sqrt{N}$ in Eq.~\eqref{eq:7}.
Nevertheless, in usual data analysis, data matrices with small dimensions are not targeted. 
For example, recent studies on FM-based black-box optimizers have involved input dimensions of approximately one hundred~\cite{fma6}.
Thus, the results in this paper are expected to be relevant for practical applications.
In Fig.~\ref{fig:result_1} at $J_1=0.1, 0.3$, the theoretical lines exhibit a plateau starting at along $\text{(coupling error)} = 5$.
These points correspond to $r$, which is determined by Eq.~\eqref{eq:8}.
The above experiments reinforce the hypothesis that our theory describes the behavior of coupling errors in the region of large $N$.

In this study, we investigated the statistics of min-max normalized eigenvalues in random matrices.
We theoretically and experimentally verified the fundamental statistical properties, such as the scaling law of a cumulative distribution and the coupling errors of matrix factorization.
For these statistics, our theory shows excellent agreement with the numerical results.

For future work, we plan to apply our analytical findings to the FM-based black-box optimizers because the lower bound of the regression error of FMs is closely related to the coupling error~\cite{rmt_fma}.
These optimizers are known for their rapid convergence of solutions and our findings may be useful for understanding this phenomenon.
Furthermore, we would like to estimate other higher-order statistics such as skewness because the use of highly nonlinear models has been proposed in recent years for black-box optimization using annealing~\cite{fm_high1,fm_high2}.

\paragraph{\footnotesize{Acknowledgments}}
\footnotesize{
We thank Yuya Seki for a lot of technical advice.
This work was partially supported by the Council for Science, Technology, and Innovation (CSTI) through the Cross-ministerial Strategic Innovation Promotion Program (SIP), ``Promoting the application of advanced quantum technology platforms to social issues'' (Funding agency: QST), Japan Science and Technology Agency (JST) (Grant Number JPMJPF2221).
One of the authors S.~T. wishes to express their gratitude to the World Premier International Research Center Initiative (WPI), MEXT, Japan, for their support of the Human Biology-Microbiome-Quantum Research Center (Bio2Q).
}

\normalsize
\appendix


\end{document}